\documentclass[a4paper]{article}

\usepackage{INTERSPEECH2020}

\usepackage{xcolor}
\usepackage[T1]{fontenc}
\usepackage[utf8]{inputenc}
\usepackage{latexsym}
\usepackage{microtype}

\usepackage{graphicx}
\usepackage{subcaption}
\usepackage{array}
\usepackage{booktabs}
\usepackage{multirow}
\usepackage{makecell}
\usepackage{longtable}
\usepackage{colortbl}
\usepackage{multicol}
\usepackage{float}
\def\myfigspace{\vspace{0mm}}

\graphicspath{{figures/}{../figures/}{../../figures}}

\newcolumntype{H}{>{\setbox0=\hbox\bgroup}c<{\egroup}@{}}
\newcolumntype{P}[1]{>{\centering\arraybackslash}p{#1}}

\def\mypar#1{\vspace{2mm}{\noindent\bfseries #1.}\hspace{1mm}}

\newcommand{\bftab}{\fontseries{b}\selectfont}

\usepackage{mathtools}
\usepackage{amssymb} % widetilde
\usepackage{mathrsfs} % mathscr
\usepackage{bbm}  % for \mathbbm 1
\usepackage{algorithm}
\usepackage{algpseudocode}

\algnewcommand{\IIf}[1]{\State\algorithmicif\ #1\ \algorithmicthen}
\algnewcommand{\EndIIf}{\unskip\ \algorithmicend\ \algorithmicif}

\algnewcommand{\LineComment}[1]{\State \(\triangleright\) #1}
\algnewcommand{\IfThenElse}[3]{% \IfThenElse{<if>}{<then>}{<else>}
  \State \algorithmicif\ #1\ \algorithmicthen\ #2\ \algorithmicelse\ #3}
  
 \algnewcommand{\IfThen}[2]{% \IfThenElse{<if>}{<then>}{<else>}
  \State \algorithmicif\ #1\ \algorithmicthen\ #2}

\usepackage{nameref}  % reference sections by name

\def\fig#1{Figure~\ref{fig:#1}}

\def\tab#1{Table~\ref{tab:#1}}

\usepackage{xspace}  % for onedot
\usepackage{enumitem}

\makeatletter
\DeclareRobustCommand\onedot{\futurelet\@let@token\@onedot}
\def\@onedot{\ifx\@let@token.\else.\null\fi\xspace}

\def\eg{\emph{e.g}\onedot} 
\def\ie{\emph{i.e}\onedot}

\def\etc{\emph{etc}\onedot}  

\def\wrt{w.r.t\onedot} 

\makeatother

\def\ind#1{[\![#1]\!]}
\newcommand{\E}[2]{\mathbb{E}_{#1}\!\left[#2\right]}
\newcommand{\cond}{\,|\,}

\newcommand{\ptt}{p_\theta}

\newcommand{\Km}{\textrm{K}}
\newcommand{\Nm}{\textrm{N}}
\newcommand{\Hm}{\textrm{H}}

\newcommand{\Hconv}{\Hm^{\text{conv}}}

\newcommand{\z}{\boldsymbol z}
\newcommand{\y}{\boldsymbol y}
\newcommand{\x}{\boldsymbol x}
\newcommand{\zw}{z^{\text{$k$}}}
\newcommand{\Zw}[1]{\boldsymbol z^{\text{$#1$}}}

\newcommand{\lx}{{|\boldsymbol x|}}
\newcommand{\ly}{{|\boldsymbol y|}}

\newcommand{\ktr}{{k_\text{train}}}
\newcommand{\kev}{{k_\text{eval}}}

\newcommand{\sto}[1][5pt]{\mathrel{%
   \hbox{\rule[\dimexpr\fontdimen22\textfont2-.2pt\relax]{#1}{.4pt}}%
   \mkern-4mu\hbox{\usefont{U}{lasy}{m}{n}\symbol{41}}}}

\definecolor{p1blue}{RGB}{0, 90, 200}     % blue
\definecolor{p1red}{RGB}{170, 10, 60}      % raspberry
\definecolor{p1green}{RGB}{10, 155, 75}    % green
\definecolor{p1yellow}{RGB}{234, 214, 68}  % yellow
\definecolor{p1orange}{RGB}{255, 130, 95}  % vermillion
\definecolor{p1purple}{RGB}{130, 20, 160}  % purple
\definecolor{p1azure}{RGB}{0, 160, 250}    % azure
\definecolor{dimgray}{rgb}{.35,.35,.35}      % rgb(35, 35, 35)

\usepackage{tikz, pgfplots}
\pgfplotsset{
    compat=1.14,
    grid style={darkgray},
    minor grid style={dimgray!20},
    major grid style={dimgray!20},
    axis line style = { darkgray }, 
    every axis plot/.append style={line width=1.5pt, mark options=solid, mark size=4pt},
    legend style={draw = darkgray, rounded corners=0pt, fill = white, font=\Large},
    tick style ={color = dimgray!30 },
    tick label style={font=\normalsize},
    label style={font=\normalsize},
}

\usetikzlibrary{decorations, shapes, arrows, arrows.meta, positioning, decorations.pathreplacing}
\usetikzlibrary{pgfplots.groupplots, patterns}
\usetikzlibrary{pgfplots.statistics}
\usepgfplotslibrary{external}

\pgfplotscreateplotcyclelist{mainpalette}{%
    smooth, gray, line width=1.2pt\\%  Offline
    smooth, black, mark=o, mark size=2pt, line width=.7pt\\%  WUE
    smooth, p1blue, mark=o, mark size=2pt, line width=.7pt\\%  k=7
    smooth, p1red, mark=o, mark size=2pt, line width=.7pt\\%  k=1
    smooth, p1green, densely dotted, mark=o, mark size=2pt, line width=.7pt\\%  ktr=kev
    smooth, p1purple, mark=o, mark size=2pt, line width=.7pt\\%  multi-k
    smooth, p1yellow, mark=o, mark size=2pt, line width=.7pt\\%  +
}

\title{Efficient Wait-$k$ Models for Simultaneous Machine Translation}

\name{Maha Elbayad\textsuperscript{1},
    Laurent Besacier\textsuperscript{1}, 
Jakob Verbeek\textsuperscript{2}} 

\address{\textsuperscript{1} Univ.\ Grenoble Alpes, CNRS, Grenoble INP, Inria, LIG, LJK, F-38000 Grenoble France\\
\textsuperscript{2} Facebook AI Research  
}

\email{\textsuperscript{1}\tt{firstname.lastname@univ-grenoble-alpes.fr}}

\begin{document}

\maketitle
 
\begin{abstract}
Simultaneous machine translation consists in starting  output generation  before the entire input sequence is available. 
Wait-$k$ decoders offer a  simple but efficient approach for this problem.
They first read $k$ source tokens, after which they  alternate between  producing a target token and reading another source token. 
We investigate the behavior of  {wait-$k$} decoding  in low resource settings for spoken corpora using  IWSLT datasets. 
We improve training of these models using unidirectional encoders, and training across multiple values of $k$. 
Experiments with Transformer and 2D-convolutional architectures show that our {wait-$k$} models generalize well across a wide range of 
latency levels.
We also show that the 2D-convolution architecture is competitive with Transformers for simultaneous translation of spoken language.
\end{abstract}

\section{Introduction}

\begin{figure}
    \centering
    \includegraphics[height=3.0cm]{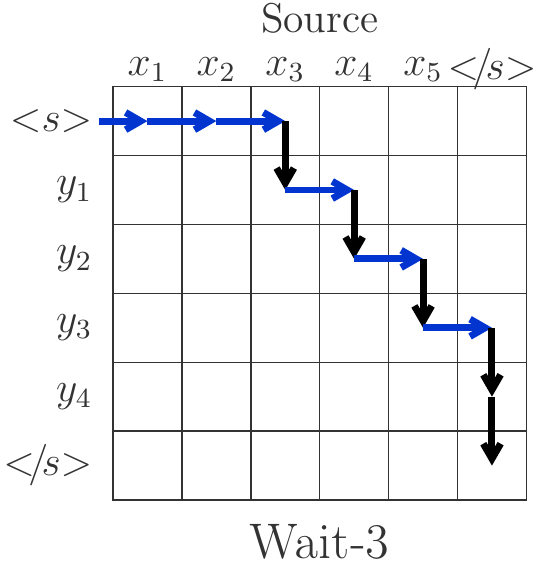}\hskip12pt
    \includegraphics[height=3.0cm]{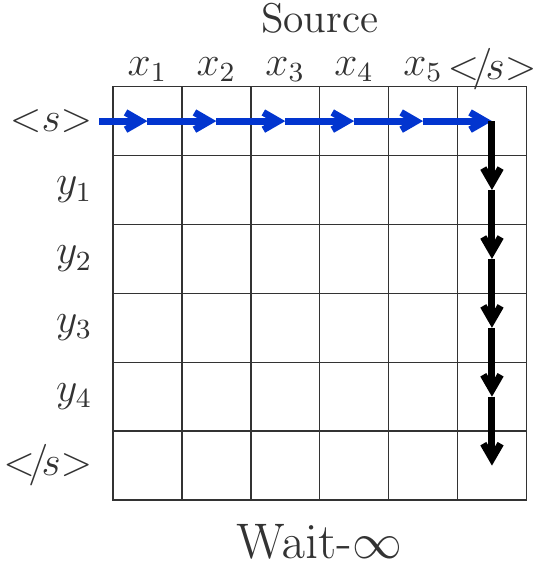}
    \vspace{-5pt}
    \caption{Wait-$k$ decoding as a sequence of reads (horizontal) and writes (vertical) over a source-target grid. After first reading $k$ tokens, the decoder alternates between reads and writes. In Wait-$\infty$, or Wait-until-End (WUE), the entire source is read first.
}\label{fig:waitk-path}
\vspace{-15pt}
\end{figure}

Neural Sequence-to-Sequence (S2S) models are state-of-the-art for sequential prediction tasks 
including  machine translation, speech recognition, speech translation, text-to-speech synthesis, \etc
The most widespread models are composed of an encoder  that reads the entire input sequence, while a decoder (often equipped with an attention mechanism) iteratively produces the next output token given the input and the partial output decoded so far.
While these models perform very well in the typical \textit{offline} decoding use case, recent works  studied  how S2S models are affected by \textit{online} (or simultaneous) constraints, and which architectures and strategies are the most efficient.
Online decoding is desirable for applications such as real-time speech-to-speech interpretation.
In such  scenarios, the decoding process starts before the entire input sequence is available, and online prediction generally comes at the cost of reduced translation quality.
In this paper we improve training and decoding of deterministic \emph{wait-$k$} models that are simple and efficient for online decoding~\cite{Ma19acl,Zheng19emnlp}.
These decoders first read $k$ tokens from the source, after which they  alternate between  producing a target token and reading another source token, see \fig{waitk-path}.

In summary our contributions are: 
(1) we propose improved training techniques for wait-$k$ by first using unidirectional encoders and training across multiple values of $k$.
(2) we show that 2D convolutional architectures are competitive with  transformers for simultaneous (online) translation, especially in lower resource settings such as encountered for  spoken corpora (IWSLT datasets), 
(3) we show that training along multiple wait-$k$ paths achieves  good online performance without the need to set a suitable $k$ \textit{a priori} for training.
Moreover, models trained in this manner generalize well across a wide range of latency levels.

\begin{figure*}
\centering
\begin{subfigure}[b]{.16\linewidth}
    \includegraphics[width=\linewidth]{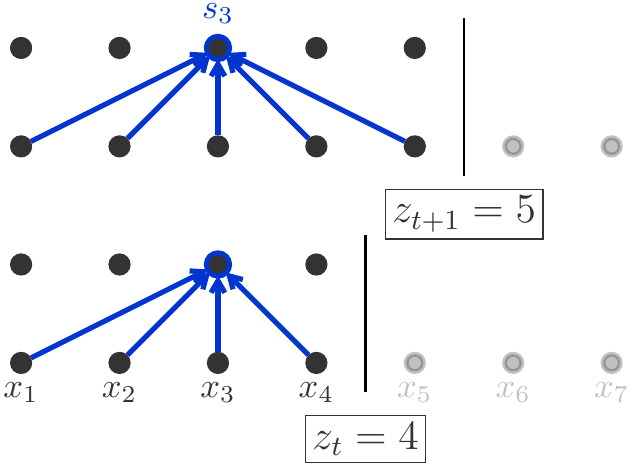}
    \caption{Bidirectional  att.}\label{fig:attention:bidir}
\end{subfigure}\hskip9pt
\begin{subfigure}[b]{.16\linewidth}
    \includegraphics[width=\linewidth]{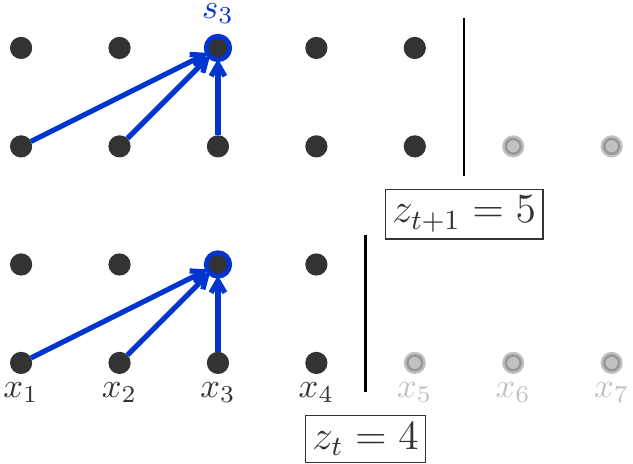}
    \caption{Unidirectional  att.}\label{fig:attention:unidir}
\end{subfigure}\hskip5pt
\begin{subfigure}[b]{.16\linewidth}
    \includegraphics[width=\textwidth]{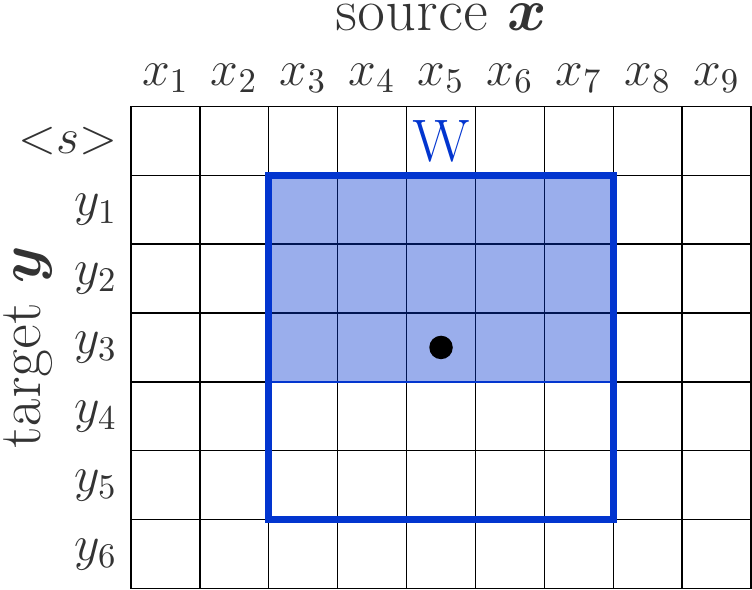}
    \caption{Offline convolution}\label{fig:offline-conv}
\end{subfigure}\hskip5pt
\begin{subfigure}[b]{.14\linewidth}
    \includegraphics[width=\textwidth]{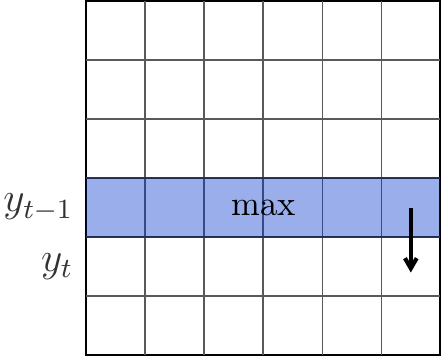}
    \caption{Offline pooling}\label{fig:offline-pool}
\end{subfigure}\hskip9pt
\begin{subfigure}[b]{.16\linewidth}
    \includegraphics[width=\textwidth]{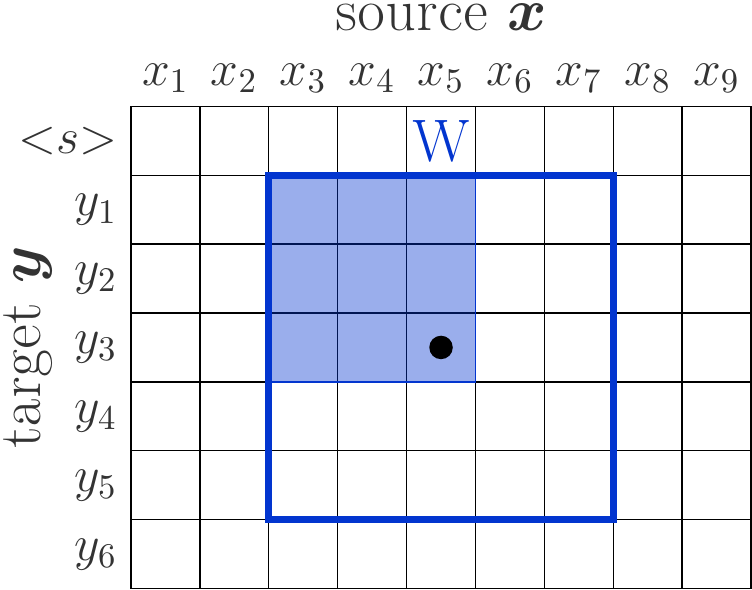}
    \caption{Online convolution}\label{fig:online-conv}
\end{subfigure}\hskip5pt
\begin{subfigure}[b]{.14\linewidth}
    \includegraphics[width=\textwidth]{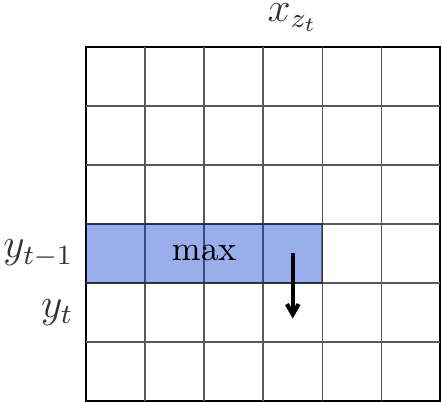}
    \caption{Online pooling}\label{fig:online-pool}
\end{subfigure}
\vspace{-5pt}
\caption{
Illustration of bi/uni-directional attention, and  causal two-dimensional convolutions. 
At the position marked with $\bullet$, the convolution only includes signals from the highlighted blue area, the other weights are zeroed out.
}
\label{fig:conv}
\vspace{-15pt}
\end{figure*}

\section{Related work}\label{sec:related}
After pioneering works on online statistical MT \cite{Fugen07MT,Yarmohammadi13naacl,He15emnlp,Grissom14emnlp,Oda15acl}, one of the first works on online translation to use attention-based sequence-to-sequence models is that of \cite{Cho16arxiv}, which uses manually designed 
 criteria that dictate whether the model should make a read/write operation.
\cite{Jaitly16neurips} reads equally-sized chunks of the source sequence and generates output sub-sequences of variable lengths, each ending with a special end-of-segment token.
\cite{Dalvi18naacl} proposed a deterministic decoding algorithm that starts with $k$ read operations then alternates between blocks of $l$ write/read operations.
This simple approach outperforms the information based criteria of \cite{Cho16arxiv}, and allows complete control of the translation delay.
\cite{Ma19acl} trained Transformer models \cite{Vaswani17nips} with a \emph{wait-$k$} decoding policy that first reads $k$ source tokens then alternate single read-writes.
Wait-$k$ approaches  were found most effective by \cite{Zheng19emnlp} when trained for the specific $k$ that is used to generate translations. 
This, however, requires training separate models for each potential value of $k$ used for translation. 

For \emph{dynamic} online decoding, \cite{Luo17icassp, Gu17eacl} rely on reinforcement learning (RL) to optimize a read/write policy.
\cite{Luo17icassp} learns an  LSTM model that emits read/write decisions based on the input and output prefixes processed so far.
\cite{Gu17eacl} trains an RNN fed with the encoder and decoder current hidden states to generate  read/write decisions. RL based models are first pre-trained offline, and then fine-tuned with policy gradient to optimize a reward balancing translation quality and latency.

To combine the end-to-end training of \emph{wait-$k$} models with the flexibility of dynamic online decoding, \cite{Press18arxiv, Zheng19acl, Zheng19emnlp} use imitation learning (IL).
\cite{Press18arxiv} estimates a decoding path from the source-target alignments obtained with an off-the-shelf alignment model then trains a recurrent network to jointly encode the two sequences following the alignment path.
\cite{Zheng19acl} adds end-of-segment tokens to the target \wrt two wait-$k$ paths during training. 
At test time, decoding is controlled with the end-of-segment token and constrained to lie in between  the two training paths.
\cite{Zheng19emnlp} supervises the training of a decoder that controls the read/write decision using  an oracle derived from an offline translation model.

Recent work on \emph{dynamic} online translation use monotonic alignments \cite{Raffel17icml}.
They were first introduced as a substitute for attention that enables progressive reading of the source context and where only a single (last read) encoder state is fed to the decoder. 
\emph{MoChA} \cite{Chiu18iclr} adds chunkwise attention on top of monotonic alignments to attend to a window of the last encoder states and \emph{MILk} \cite{Arivazhagan19acl} broadens this window with an infinite lookback to boost the translation quality.
\cite{Ma20iclr} adapted \emph{MILk}'s monotonic attention for multi-headed Transformer decoders.

Simultaneous translation models usually operate under a \emph{streaming} constraint where an emitted output cannot be altered, alternatively, \cite{Niehues16interspeech, Arivazhagan20iwslt, Zheng20acl} propose decoding strategies that allow for revision to correct past outputs.

In our work, we focus on wait-$k$ decoding,
but unlike \cite{Ma19acl} we opt for unidirectional encoders, and show that they are more effective and efficient  for online translation. 
We also show that it is possible to train a single model that is effective across a large range of latency levels.

\section{Online translation  models}\label{sec:models}

In the following we describe how we adapt the transformer \cite{Vaswani17nips} and pervasive attention  \cite{Elbayad18conll} architectures to the online translation setting, and how we train them for wait-$k$ decoding. 

\subsection{Online Transformer (TF)}\label{sec:transformer}

The key component of the transformer model \cite{Vaswani17nips} is multi-headed attention, which concatenates  the outputs of multiple attention heads.
Attention aggregates encoder/decoder states from other positions in the form of a weighted sum, where the weights depend on the current state.

Wait-$k$ online decoding~\cite{Ma19acl,Dalvi18naacl}
 starts by reading $k$ source tokens, and  then alternates between reading and writing a single token at a time, until  the full source has been read, or the target generation has been terminated.
Formally, we denote with $z_t$ the number of source tokens read when decoding $y_t$. For a wait-$k$ decoding path we have  $z_t = \min(k+t-1, \lx).$ 

In the encoder-decoder attention, the decoder state $h_{t}$, used to predict $y_{t+1}$,  attends to the first $z_{t+1}$ source states, and each  source state should only encode signal from the $z_t$ source tokens read so far. 
Self-attention makes the  source encoder bidirectional,  \ie the encoder state at a given position includes signals from past as well as future time-steps.
This means that, as in \cite{Ma19acl,Zheng19emnlp},  
the bidirectional source encoding has to be recomputed each time a source token is read, 
so that  past source tokens  can attend to the newly available source tokens, see \fig{attention:bidir}. 

To alleviate the cost of re-encoding the input sequence after each read operation,
we propose unidirectional encoders for online translation, 
by masking the self-attention to only consider previous time-steps, as in \fig{attention:unidir}.
In this manner, 
unlike~\cite{Ma19acl,Zheng19emnlp},
both during training and deployment,  source sequences are  encoded once, without having to update the encoder states as new source tokens  become available. 

\subsection{Online Pervasive Attention (PA)}

In Pervasive Attention~\cite{Elbayad18conll}, the source and target sequences are jointly encoded with a two-dimensional convolutional neural network (CNN).
For  decoding, \cite{Elbayad18conll} uses causal convolutions where the filters mask future target positions, see \fig{offline-conv}. 
In a similar way, to adapt to the online translation task, we mask the filters in the source direction in order to encode the source unidirectionally, see \fig{online-conv}.
In our online version of Pervasive Attention, the CNN's ultimate features $\Hconv$ at a given position $(t,j)$ of the source-target grid encode source context up to $x_j$ and target context up to $y_t$.
In the  offline Pervasive Attention, a single representation per target position is needed to predict the next output token, which is obtained using max-pooling across the source positions, see \fig{offline-pool}. 
In the online task, we would like to make a prediction at every position where the model could be asked to write.
When predicting $y_t$ at position $(t-1,z_t)$ of the grid we max-pool the activations in $\Hconv_{t-1, \leq z_t}$, see \fig{online-pool}. 

One major difference between Pervasive Attention (PA) and Transformer is that in PA, the source-target representation at any given position $(t,z_t)$ is independent from the decoding path $\z$ taken to get to that point.
In a transformer model, however, the representation at $(t,z_t)$ depends on the order in which tokens were read and written up to that point.

\subsection{Training wait-$k$ models}

In \cite{Ma19acl,Dalvi18naacl} the model is optimized by maximum likelihood estimation \wrt a single wait-$k$ decoding path $\Zw k$:
\begin{align}
\log p(\y\cond\x,\Zw k) = 
\sum_{t=1}^\ly \log p_\theta(y_t|\y_{<t},\x_{\leq \zw_t},\Zw{k}_{<t}).
\end{align}
Note that the dependency on $\Zw{k}_{<t}$ is only relevant for the Transformer model where the path history matters.

Instead of optimizing a single decoding path, we propose to jointly optimize across multiple wait-$k$ paths. 
The additional loss terms provide a richer training signal, and potentially yield models that could perform well in different latency regimes. 
Due to the dependence of the decoder hidden states on the full decoding path $\z_{<t}$ in the transformer-based model, 
we can only train in parallel across a limited set of paths. 
We consider wait-$k$ paths $\{\Zw{k},\forall k \in \Km\}$. 
During training, we encode the source sequence once, and uniformly sample $k\in\Km$ to decode:
\begin{align}
   \E{\Km}{\log p(\y\cond\x,\Zw k)} 
   \approx 
   \sum_{k\sim \mathcal U(\Km)} \log p_\theta(\y | \x,\Zw k).
\end{align}
To cover all possible wait-$k$ paths for an input $(\x,\y)$, we set $\Km=[1,\dots,\lx]$.
We will refer to this training with \emph{multi-path}. 

With Pervasive Attention, we can leverage more training signals.
In fact, the grid nature of the model allows us to efficiently compute the output distributions $p(y_t|\y_{<t},\x_{\leq j})$ all over the grid in a single forward pass.
Consequently, we optimize the writing log-likelihoods in the area above the diagonal with:
\begin{align}
 \sum_{t=1}^\ly \sum_{j=1}^\lx \log \ptt(y_t | \y_{<t}, \x_{\leq j}) \ind{j\geq t}.
\end{align}
We will refer to this training with \emph{multi-path} as well.

\section{Experimental evaluation}\label{sec:exp}
\subsection{Datasets and experimental setup}

\begin{table} 
\centering
\caption{Evaluation of  Pervasive Attention (PA) and Transformer (TF) for offline translation with  greedy decoding.
}
\vspace{-5pt}
\begin{tabular}{cccHH|ccHH}
\toprule
Encoder & \multicolumn{4}{c}{Unidirectional} &
\multicolumn{4}{|c}{Bidirectional} \\
\midrule
Architecture & PA & TF & PA & TF & PA & TF & PA & TF \\
\midrule
IWSLT'14 En$\sto$De & \bftab 26.81 & 26.58 
                   & 27.54 & 27.70 
                   & 27.23 & \bftab27.46 
                   & 27.96 & \bftab28.27\\
IWSLT'14 De$\sto$En & 32.40 & \bftab32.81 
                   & 33.62 & 33.79 
                   & 33.43 & \bftab33.64 
                   & 34.36 & \bftab34.56 \\
\midrule
IWSLT'15 En$\sto$Vi &  \bftab29.22 & 28.90 
                   & 30.20 & 29.65 
                   & \bftab29.81 & 29.33 
                   & \bftab30.49 & 30.17\\
IWSLT'15 Vi$\sto$En & \bftab26.81 & 25.73 
                   & 27.96 & 26.92 
                   & 27.43 & \bftab28.09 
                   & 28.04 &  \bftab29.00\\
\midrule
WMT'15 De$\sto$En   & 28.08 & \bftab31.14 
                   & 29.28 & 32.27 
                   & 28.78 &  \bftab31.96
                   & 29.61 & \bftab32.80\\
\bottomrule
\end{tabular}

\label{tab:baselines}
\vspace{-10pt}
\end{table}

We evaluate our approach on IWSLT14 En$\leftrightarrow$De \cite{Cettolo14iwslt}, IWSLT'15 En$\leftrightarrow$Vi \cite{Luong15iwslt}, and WMT15 De$\sto$En datasets.\footnote{http://www.statmt.org/wmt15/} We train offline unidirectional and bidirectional Transformer (TF) and Pervasive Attention  (PA) models on all tasks. 
On IWSLT'14 De$\leftrightarrow$En, similar to \cite{Elbayad18conll,Edunov18naacl}, we train on 160K pairs, develop on 7K held out pairs and test on TED dev2010+tst2010-2013 (6,750 pairs).
All data is tokenized and lower-cased and we segment sequences using byte pair encoding \cite{Sennrich16acl} with 10K merge operations. The resulting vocabularies are of 8.8K and 6.6K types in German and English respectively. 
On IWSLT'15 En$\leftrightarrow$Vi, similar to~\cite{Ma20iclr,Luong15iwslt}, we train on 133K pairs, develop on TED tst2012 (1,553 pairs) and test on TED tst2013 (1,268 pairs). 
The corpus was simply tokenized resulting in 17K and 7.7K word vocabularies in English and Vietnamese respectively.
On WMT'15 De$\sto$En, we reproduce the setup of \cite{Ma19acl, Arivazhagan19acl} with a joint vocabulary of 32K BPE types. We train on 4.5M pairs, develop on newstest2013 (3,000 pairs) and test on newstest15 (2,169 pairs).

Our Pervasive Attention models use residual-cumulative skip connections and stack $\Nm=14$ layers with $11\times 11$ convolutions. 
We train Transformer \emph{small} on IWSLT'14 De$\leftrightarrow$En, a modified \emph{base} \cite{Ma20iclr} on IWSLT'15 En$\leftrightarrow$Vi and Transformers \emph{base} and \emph{big} on WMT'15 De$\sto$En.

We evaluate all models with greedy decoding, and measure translation quality  measured with tokenized word-level BLEU~\cite{papineni02acl} with \emph{multi-bleu.pl}. 
We measure decoding latency with Average Lagging (AL) \cite{Ma19acl}, which  is designed to indicate the source steps by which we lag behind an ideal translator (wait-0), it can however be negative if the system finishes decoding prematurely before the full source is read.
Other measures of lagging include \emph{Average proportion} (AP) \cite{Cho16arxiv} and \emph{Differentiable Average Lagging} (DAL) \cite{Arivazhagan19acl}.
AP is unfavorable to short sequences and is incapable of highlighting improvement as it occupies a narrow range~\cite{Ma19acl,Arivazhagan19acl,Alinejad18emnlp}.
DAL is a differentiable version of AL used to regularize trainable decoders, and behaves similarly to AL.

\subsection{Offline comparison}
 \tab{baselines} reports offline performance of Pervasive Attention (PA) and Transformer (TF) models with both a unidirectional encoder and a bidirectional encoder.
Overall, and as expected, bidirectional encoders in the offline setup are better than their unidirectional counterparts. 
The gain for PA is of 0.65 on average while for  TF the addition of bidirectionality improves BLEU by 1.1 on average. 
The first two columns of \tab{baselines} show that pervasive attention (PA) is  competitive with TF on these  datasets when using unidirectional encoders: PA improves upon TF on three of the five  tasks. 

\subsection{Online comparison}\label{sec:online-comparison}

\begin{figure}
\centering
\hspace{8pt}\begin{tikzpicture}[scale=0.7]
   \begin{axis}[%
    hide axis,
    scale only axis,
    xmin=-2, xmax=-1,
    ymin=0, ymax=1,
    legend style={legend cell align=left, font=\small, column sep=1pt},
    legend columns=4,
    ]

    \addlegendimage{mark=x}
    \addlegendentry{Bidirectional\phantom{a}}
    \addlegendimage{mark=o}
    \addlegendentry{Unidirectional\phantom{a}}
    \addlegendimage{p1red}
    \addlegendentry{$\ktr=1$}
    \addlegendimage{p1blue}
    \addlegendentry{$\ktr=7$}
    \end{axis}
\end{tikzpicture}
\vspace{-3pt}
\begin{subfigure}[b]{.5\linewidth}
    \def\results{data/iwslt_deen}
\begin{tikzpicture}
    \pgfmathsetmacro{\avleng}{22.97}
    \pgfmathsetmacro{\wue}{32.21}
    \begin{axis}[height=3.8cm, width=4.8cm, grid=both,
                 y axis line style=-,
                 xlabel=\csname AL\endcsname, 
                 x label style={at={(axis description cs:0.5,-0.07)},anchor=north, font=\scriptsize},
                 ylabel=BLEU,
                 y label style={font=\scriptsize},
                 y tick label style={font=\tiny, xshift=2pt},
                 x tick label style={font=\tiny, yshift=1pt},
                 xmin=-2,xmax=9, 
                 xtick={-1,1,3,5,7,9},
                 minor x tick num=1,
                 ymin=15, ymax=32,
                 ytick={13,15,...,33},
                 minor y tick num=1,
                 legend style={at={(0,1)}, anchor=north west, font=\scriptsize},
                 every axis plot/.append style={line width=0.8pt, mark size=2pt},
                 ]

\addplot[p1red, mark=x]
table [y=BLEU4-b1,x=AL-b1]{\results/bidir_transformer/trained_path_k1.dat};

\addplot[p1red, mark=o]
table [y=BLEU4-b1,x=AL-b1]{\results/unidir_transformer/trained_path_k1.dat};

\addplot[p1blue, mark=x]
table [y=BLEU4-b1,x=AL-b1]{\results/bidir_transformer/trained_path_k7.dat};

\addplot[p1blue, mark=o]
table [y=BLEU4-b1,x=AL-b1]{\results/unidir_transformer/trained_path_k7.dat};

\end{axis}
\end{tikzpicture}
    \caption{IWSLT'14 De$\sto$En - TF}
\end{subfigure}%
\begin{subfigure}[b]{.5\linewidth}
    \def\results{data/iwslt15_envi}
\begin{tikzpicture}
    \pgfmathsetmacro{\avleng}{21.08}
    \pgfmathsetmacro{\wue}{28.90}
    \begin{axis}[height=3.8cm, width=4.8cm, grid=both,
                 y axis line style=-,
                 xlabel=\csname AL\endcsname, 
                 x label style={at={(axis description cs:0.5,-0.07)},anchor=north, font=\scriptsize},
                 %ylabel=BLEU,
                 y tick label style={font=\tiny, xshift=2pt},
                 x tick label style={font=\tiny, yshift=1pt},
                 xmin=2,xmax=10.1, 
                 xtick={-1,1,3,5,7,9,11},
                 minor x tick num=1,
                 ymin=18, ymax=30,
                 ytick={18,20,...,30},
                 minor y tick num=1,
                 legend style={anchor=north east, font=\scriptsize},
                 every axis plot/.append style={line width=0.8pt, mark size=2pt},
                 ]

\addplot[mark=x, p1red]
table [y=BLEU4-b1,x=AL-b1]{\results/bidir_transformer/trained_path_k1.dat};
\addplot[mark=o, p1red]
table [y=BLEU4-b1,x=AL-b1]{\results/unidir_transformer/trained_path_k1.dat};

\addplot[mark=x, p1blue]
table [y=BLEU4-b1,x=AL-b1]{\results/bidir_transformer/trained_path_k7.dat};
\addplot[mark=o, p1blue]
table [y=BLEU4-b1,x=AL-b1]{\results/unidir_transformer/trained_path_k7.dat};
\end{axis}
\end{tikzpicture}
    \caption{IWSLT'15 En$\sto$Vi - TF}
\end{subfigure}
\vspace{-15pt}
\caption{
Transformer models with bi/uni-directional encoders trained on  wait-1 and wait-7 decoding  paths.
}
\label{fig:direction}
\vspace{-10pt}
\end{figure}
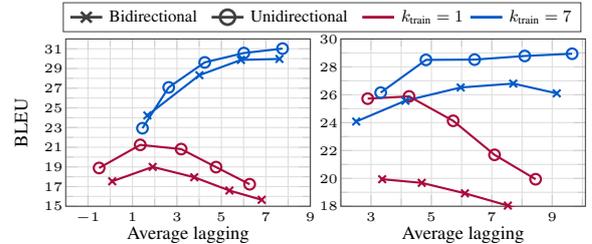

For the Transformer model, 
we initially consider a bidirectional encoder similar to \cite{Ma19acl, Zheng19emnlp}, in which case the encoder states have to be updated after each read. 
The timing results in \tab{timing} show that using a bidirectional encoder rather than a unidirectional one slows  decoding down by a factor  two to three.
Second, in \fig{direction}
we assess the impact of the uni/bidirectional encoder on  online decoding quality. 
We look at models trained using either of two wait-$k$ paths: $\ktr{=}1$ and $\ktr{=}7$. 
We observe that in the case of online decoding  unidirectional encoding performs best, in contrast to the case for offline decoding. 
For both  $\ktr{=}1$ and $\ktr{=}7$, the  unidirectional encoder  is consistently providing better performance. 
For the experiments below we therefore use unidirectional encoders. 

\begin{table}
\centering
\caption{Decoding speed of  Transformers  with uni/bi-directional encoders for De$\sto$En  on IWSLT'14 and WMT'15.
}
\vspace{-5pt}
\begin{tabular}{rlHrr}
\toprule
& \multicolumn{4}{r}{Decoding (tok/s)} \\
& Encoder  &  & GPU & CPU \\
\midrule
\multirow{2}{*}{\shortstack{IWSLT\\De$\sto$En}}
& Unidirectional & 2.9 & 21.7k & 130 \\
& Bidirectional&  116 & 7.3k & 54  \\
\midrule
\multirow{2}{*}{\shortstack{WMT\\De$\sto$En}}
& Unidirectional & 389 & 6.3k &  77  \\
& Bidirectional  &   -         & 2.9k &  32  \\
\bottomrule
\end{tabular}

\label{tab:timing}
\myfigspace
\end{table}

\mypar{Pervasive Attention and Transformer for online translation}
We evaluate  models  trained for different wait-$k$ decoding paths.
We denote with $\ktr{=}\infty$ the wait-until-end training where the full source is read before decoding. 
We report offline results for reference, the offline model has a latency of $\textrm{AL}=\lx$.

\begin{figure*}
\centering
\begin{tikzpicture}[scale=0.8]
   \begin{axis}[%
    hide axis,
    scale only axis,width=1mm,
    xmin=-2, xmax=-1,
    ymin=0, ymax=1,
    legend style={legend cell align=left, font=\small, column sep=1pt},
    legend columns=6,
    cycle list name=mainpalette,
    ]
    \pgfplotsinvokeforeach{1,...,6}{\addplot coordinates {(0,0)};} % make six dummy plots

    \addlegendentry{Offline\phantom{a}}
    \addlegendentry{$\ktr=\infty$\phantom{a}}
    \addlegendentry{$\ktr=7$\phantom{a}}
    \addlegendentry{$\ktr=1$\phantom{a}}
    \addlegendentry{$\ktr=\kev$\phantom{a}}
    \addlegendentry{\emph{multi-path}}
    \end{axis}
\end{tikzpicture}
\begin{subfigure}[b]{.23\linewidth}
    \def\results{data/iwslt_deen}
\begin{tikzpicture}
\pgfplotsset{compat=1.11,
    legend image code/.code={
        \draw[mark repeat=2,mark phase=2, mark size=1.5pt, line width=0.7pt]
            plot coordinates {
                (0cm,0cm)
                (0.15cm,0cm)        %% default is (0.3cm,0cm)
                (0.3cm,0cm)         %% default is (0.6cm,0cm)
            };}}

\pgfmathsetmacro{\avleng}{22.97}
\pgfmathsetmacro{\wue}{32.40}
\begin{groupplot}[
    group style={%
        group size=2 by 1,
        horizontal sep=0pt
    },
    ymin=16, ymax=33.4,
    legend style={at={(1.18,0.04)},anchor=south east, font=\tiny, inner sep=1pt},
    tick label style={font=\tiny},
    y tick label style={font=\tiny, xshift=2pt},
    x tick label style={font=\tiny, yshift=1pt},
    ytick={16,20,...,32},
    minor y tick num=1,
   ]
   \nextgroupplot[xmin=-2,xmax=9,
                   height=3.8cm,
                   width=4.5cm, grid=both,
                   y axis line style=-,
                   xlabel=\csname AL\endcsname, 
                   x label style={at={(axis description cs:0.5,-0.07)},anchor=north, font=\scriptsize},
                   ylabel=BLEU,
                   y label style={font=\scriptsize},
                   xtick={-1,1,3,5,7,9},
                   minor x tick num=1,
                   cycle list name=mainpalette,
                   legend to name=DeEnMPA,
                   ]

\addplot coordinates {(-3, \wue) (22, \wue)};
\addlegendentry{Offline}

\addplot table [y=BLEU4-b1,x=AL-b1]{\results/attn2d/trained_path_kinf_update.dat};
\addlegendentryexpanded{$\ktr=\infty$}

\addplot table [y=BLEU4-b1,x=AL-b1]{\results/attn2d/trained_path_k7_update.dat};
\addlegendentry{$\ktr=7$}

\addplot table [y=BLEU4-b1,x=AL-b1]{\results/attn2d/trained_path_k1_update.dat};
\addlegendentry{$\ktr=1$}

\addplot table [y=BLEU4-b1,x=AL-b1]{\results/attn2d/keval_update.dat};
\addlegendentry{$\ktr=\kev$}

\addplot table [y=BLEU4-b1,x=AL-b1]{\results/attn2d/above_n14k11_max.dat};
\addlegendentry{$\ktr\geq 1$}

%\legend{}
\end{groupplot}
%\node[anchor=south east] at (rel axis cs: 1.03,-0.03) {\pgfplotslegendfromname{DeEnMPA}};

\end{tikzpicture}
    \caption{De$\sto$En Pervasive Attention}\label{fig:deen-mpa-path}
\end{subfigure}\hskip1pt
\begin{subfigure}[b]{.21\linewidth}
    \def\results{data/iwslt_ende}
\begin{tikzpicture}
    \pgfmathsetmacro{\avleng}{22.21}
    \pgfmathsetmacro{\wue}{26.81}

    \begin{groupplot}[
        group style={%
            group size=2 by 1,
            horizontal sep=0pt
        },
        ymin=13, ymax=28,
        legend style={at={(1.2,0.04)},anchor=south east},
        tick label style={font=\tiny},
        y tick label style={font=\tiny, xshift=2pt},
        x tick label style={font=\tiny, yshift=1pt},
        ytick={13,16,...,32},
        minor y tick num=1,
       ]
    \nextgroupplot[xmin=0.5,xmax=9,
                   height=3.8cm,
                   width=4.5cm, grid=both,
                   xlabel=\csname AL\endcsname, 
                   x label style={at={(axis description cs:0.5,-0.07)},anchor=north, font=\scriptsize},
                   %ylabel=BLEU,
                   xtick={-1,1,3,5,7,9},
                   minor x tick num=1,
                   cycle list name=mainpalette,
                   ]

\addplot coordinates {(-1, \wue) (22, \wue)};
\addlegendentry{Offline}

\addplot table [y=BLEU4-b1,x=AL-b1]{\results/attn2d/trained_path_kinf_update.dat};
\addlegendentryexpanded{$\ktr=\infty$}

\addplot table [y=BLEU4-b1,x=AL-b1]{\results/attn2d/trained_path_k7_update.dat};
\addlegendentry{$\ktr=7$}

\addplot table [y=BLEU4-b1,x=AL-b1]{\results/attn2d/trained_path_k1_update.dat};
\addlegendentry{$\ktr=1$}

\addplot table [y=BLEU4-b1,x=AL-b1]{\results/attn2d/keval_update.dat};
\addlegendentry{$\ktr=\kev$}

\addplot table [y=BLEU4-b1,x=AL-b1]{\results/attn2d/trained_above_k1_update.dat};
\addlegendentry{$k\geq k_\text{eval}$}

\legend{}

\end{groupplot}
\end{tikzpicture}
    \caption{En$\sto$De Pervasive Attention}\label{fig:ende-mpa-path}
\end{subfigure}\hskip1pt
\begin{subfigure}[b]{.21\textwidth}
    \def\results{data/iwslt15_envi}
\begin{tikzpicture}
    \pgfmathsetmacro{\avleng}{21.08}
    \pgfmathsetmacro{\wue}{29.22}

    \begin{groupplot}[
        group style={%
            group size=2 by 1,
            horizontal sep=0pt
        },
        ymin=24, ymax=30,
        legend style={at={(1.2,0.04)},anchor=south east},
        tick label style={font=\tiny},
        y tick label style={font=\tiny, xshift=2pt},
        x tick label style={font=\tiny, yshift=1pt},
        ytick={14,15,...,32},
        minor y tick num=1,
       ]
    \nextgroupplot[xmin=2,xmax=10,
                   height=3.8cm,
                   width=4.5cm, grid=both,
                   y axis line style=-,
                   xlabel=\csname AL\endcsname, 
                   x label style={%
                   at={(axis description cs:0.5,-0.07)},anchor=north, font=\scriptsize},
                   %ylabel=BLEU,
                   xtick={0,2,...,10},
                   minor x tick num=1,
                   cycle list name=mainpalette,
                   ]

\addplot coordinates {(-1, \wue) (22, \wue)};
\addlegendentry{Offline}

\addplot table [y=BLEU4-b1,x=AL-b1]{\results/attn2d/trained_path_kinf_update.dat};
\addlegendentryexpanded{$\ktr=\infty$}

\addplot table [y=BLEU4-b1,x=AL-b1]{\results/attn2d/trained_path_k7_update.dat};
\addlegendentry{$\ktr=7$}

\addplot table [y=BLEU4-b1,x=AL-b1]{\results/attn2d/trained_path_k1_update.dat};
\addlegendentry{$\ktr=1$}

\addplot table [y=BLEU4-b1,x=AL-b1]{\results/attn2d/keval_update.dat};
\addlegendentry{$\ktr=\kev$}

\addplot table [y=BLEU4-b1,x=AL-b1]{\results/attn2d/trained_above_k1_update.dat};
\addlegendentry{$\ktr\geq1$}

\legend{}
\end{groupplot}
\end{tikzpicture}
    \caption{En$\sto$Vi Pervasive Attention}\label{fig:envi-mpa-path}
\end{subfigure}\hskip1pt
\begin{subfigure}[b]{.21\textwidth}
    \def\results{data/iwslt15_vien}
\begin{tikzpicture}
    \pgfmathsetmacro{\avleng}{26.56}
    \pgfmathsetmacro{\wue}{26.81}

    \begin{groupplot}[
        group style={%
            group size=2 by 1,
            horizontal sep=0pt
        },
        ymin=8, ymax=28,
        legend style={at={(1.2,0.04)},anchor=south east},
        tick label style={font=\tiny},
        y tick label style={font=\tiny, xshift=2pt},
        x tick label style={font=\tiny, yshift=1pt},
        ytick={8,12,...,28},
        minor y tick num=1,
       ]
    \nextgroupplot[xmin=-4.1,xmax=8,
                   height=3.8cm,
                   width=4.5cm, grid=both,
                   y axis line style=-,
                   xlabel=\csname AL\endcsname, 
                   x label style={%
                   at={(axis description cs:0.5,-0.07)},anchor=north, font=\scriptsize},
                   %ylabel=BLEU,
                   y label style={font=\scriptsize},
                   xtick={-3,-1,1,3,...,10},
                   minor x tick num=1,
                   cycle list name=mainpalette,
                   ]

\addplot  coordinates {(-5, \wue) (22, \wue)};
\addlegendentry{Offline}

\addplot table [y=BLEU4-b1,x=AL-b1]{\results/attn2d/trained_path_kinf_update.dat};
\addlegendentryexpanded{$\ktr=\infty$}

\addplot table [y=BLEU4-b1,x=AL-b1]{\results/attn2d/trained_path_k7_update.dat};
\addlegendentry{$\ktr=7$}

\addplot table [y=BLEU4-b1,x=AL-b1]{\results/attn2d/trained_path_k1_update.dat};
\addlegendentry{$\ktr=1$}

\addplot table [y=BLEU4-b1,x=AL-b1]{\results/attn2d/keval_update.dat};
\addlegendentry{$\ktr=\kev$}

\addplot table [y=BLEU4-b1,x=AL-b1]{\results/attn2d/trained_above_k1_update.dat};
\addlegendentry{$\ktr\geq1$}

\legend{}
\end{groupplot}
\end{tikzpicture}
    \caption{Vi$\sto$En Pervasive Attention}\label{fig:vien-mpa-path}
\end{subfigure}\\
\begin{subfigure}[b]{.23\linewidth}
    \def\results{data/iwslt_deen}
\begin{tikzpicture}
    \pgfmathsetmacro{\avleng}{22.97}
    \pgfmathsetmacro{\wue}{32.81}
    \begin{groupplot}[
        group style={%
            group size=2 by 1,
            horizontal sep=0pt
        },
        ymin=16, ymax=33.4,
        legend style={at={(1.2,0.04)},anchor=south east},
        tick label style={font=\tiny},
        y tick label style={font=\tiny, xshift=2pt},
        x tick label style={font=\tiny, yshift=1pt},
        ytick={16,20,...,32},
        minor y tick num=1,
       ]
    \nextgroupplot[xmin=-2,xmax=9,
                   height=3.8cm,
                   width=4.5cm, grid=both,
                   y axis line style=-,
                   xlabel=\csname AL\endcsname, 
                   x label style={at={(axis description cs:0.5,-0.07)},anchor=north, font=\scriptsize},
                   ylabel=BLEU,
                   y label style={font=\scriptsize},
                   xtick={-1,1,3,5,7,9},
                   minor x tick num=1,
                   cycle list name=mainpalette,
                   ]
    
\addplot coordinates {(-3, \wue) (22, \wue)};
\addlegendentry{Offline}

\addplot table [y=BLEU4-b1,x=AL-b1]{\results/unidir_transformer/trained_path_kinf.dat};
\addlegendentryexpanded{$\ktr=\infty$}

\addplot table [y=BLEU4-b1,x=AL-b1]{\results/unidir_transformer/trained_path_k7.dat};
\addlegendentry{$\ktr=7$}

\addplot table [y=BLEU4-b1,x=AL-b1]{\results/unidir_transformer/trained_path_k1.dat};
\addlegendentry{$\ktr=1$}

\addplot table [y=BLEU4-b1,x=AL-b1]{\results/unidir_transformer/keval.dat};
\addlegendentry{$\ktr=\kev$}

\addplot table [y=BLEU4-b1,x=AL-b1]{\results/unidir_transformer/multipath_1tolx.dat};
\addlegendentry{$\ktr\in[1..\lx]$}

\legend{}
\end{groupplot}
\end{tikzpicture}
    \caption{De$\sto$En Transformer}\label{fig:deen-mt-path}
\end{subfigure}\hskip1pt
\begin{subfigure}[b]{.21\linewidth}
    \def\results{data/iwslt_ende}
\begin{tikzpicture}
    \pgfmathsetmacro{\avleng}{22.21}
    \pgfmathsetmacro{\wue}{26.58}
    \begin{groupplot}[
        group style={%
            group size=2 by 1,
            horizontal sep=0pt
        },
        ymin=13, ymax=28,
        legend style={at={(1.2,0.04)},anchor=south east, 
            font=\tiny, 
            legend cell align=left},
        tick label style={font=\tiny},
        y tick label style={font=\tiny, xshift=2pt},
        x tick label style={font=\tiny, yshift=1pt},
        ytick={13,16,...,32},
        minor y tick num=1,
       ]
    \nextgroupplot[xmin=1,xmax=9,
                   height=3.8cm,
                   width=4.5cm, grid=both,
                   y axis line style=-,
                   xlabel=\csname AL\endcsname, 
                   x label style={at={(axis description cs:0.5,-0.07)},anchor=north, 
                   font=\scriptsize},
                   %ylabel=BLEU,
                   xtick={-1,1,3,5,7,9},
                   minor x tick num=1,
                   cycle list name=mainpalette,
                   ]
    
\addplot coordinates {(-1, \wue) (22, \wue)};
\addlegendentry{Offline}

\addplot table [y=BLEU4-b1,x=AL-b1]{\results/unidir_transformer/trained_path_kinf.dat};
\addlegendentryexpanded{$\ktr=\infty$}

\addplot table [y=BLEU4-b1,x=AL-b1]{\results/unidir_transformer/trained_path_k7.dat};
\addlegendentry{$\ktr=7$}

\addplot table [y=BLEU4-b1,x=AL-b1]{\results/unidir_transformer/trained_path_k1.dat};
\addlegendentry{$\ktr=1$}

\addplot table [y=BLEU4-b1,x=AL-b1]{\results/unidir_transformer/keval.dat};
\addlegendentry{$\ktr=\kev$}

\addplot table [y=BLEU4-b1,x=AL-b1]{\results/unidir_transformer/1tolx.dat};
\addlegendentry{$\ktr\in[1..\lx]$}

\legend{}

\end{groupplot}
\end{tikzpicture}
    \caption{En$\sto$De Transformer}\label{fig:ende-mt-path}
\end{subfigure}\hskip1pt
\begin{subfigure}[b]{.21\textwidth}
    \def\results{data/iwslt15_envi}
\begin{tikzpicture}
    \pgfmathsetmacro{\avleng}{21.08}
    \pgfmathsetmacro{\wue}{28.90}

    \begin{groupplot}[
        group style={%
            group size=2 by 1,
            horizontal sep=0pt
        },
        ymin=24, ymax=30,
        legend style={at={(1.2,0.04)},anchor=south east},
        tick label style={font=\tiny},
        y tick label style={font=\tiny, xshift=2pt},
        x tick label style={font=\tiny, yshift=1pt},
        ytick={14,15,...,32},
        minor y tick num=1,
       ]
    \nextgroupplot[xmin=2,xmax=10,
                   y axis line style=-,
                   height=3.8cm,
                   width=4.5cm, grid=both,
                   xlabel=\csname AL\endcsname, 
                   x label style={%
                   at={(axis description cs:0.5,-0.07)},anchor=north, font=\scriptsize},
                   %ylabel=BLEU,
                   xtick={0,2,...,10},
                   minor x tick num=1,
                   cycle list name=mainpalette,
                   ]

\addplot  coordinates {(-1, \wue) (22, \wue)};
\addlegendentry{Offline}

\addplot table [y=BLEU4-b1,x=AL-b1]{\results/unidir_transformer/trained_path_kinf.dat};
\addlegendentryexpanded{$\ktr=\infty$}

\addplot table [y=BLEU4-b1,x=AL-b1]{\results/unidir_transformer/trained_path_k7.dat};
\addlegendentry{$\ktr=7$}

\addplot table [y=BLEU4-b1,x=AL-b1]{\results/unidir_transformer/trained_path_k1.dat};
\addlegendentry{$\ktr=1$}

\addplot table [y=BLEU4-b1,x=AL-b1]{\results/unidir_transformer/keval.dat};
\addlegendentry{$\ktr=\kev$}

\addplot table [y=BLEU4-b1,x=AL-b1]{\results/unidir_transformer/1tolx.dat};
\addlegendentry{$\ktr\in[1..\lx]$}

\legend{}

\end{groupplot}
\end{tikzpicture}
    \caption{En$\sto$Vi Transformer}\label{fig:envi-mt-path}
\end{subfigure}\hskip1pt
\begin{subfigure}[b]{.21\textwidth}
    \def\results{data/iwslt15_vien}
\begin{tikzpicture}
    \pgfmathsetmacro{\avleng}{26.56}
    \pgfmathsetmacro{\wue}{25.73}

    \begin{groupplot}[
        group style={%
            group size=2 by 1,
            horizontal sep=0pt
        },
        ymin=8, ymax=28,
        legend style={at={(1.2,0.04)},anchor=south east},
        tick label style={font=\tiny},
        y tick label style={font=\tiny, xshift=2pt},
        x tick label style={font=\tiny, yshift=1pt},
        ytick={8,12,...,28},
        minor y tick num=1,
       ]
    \nextgroupplot[xmin=-4.1,xmax=8,
                   height=3.8cm,
                   width=4.5cm, grid=both,
                   y axis line style=-,
                   xlabel=\csname AL\endcsname, 
                   x label style={%
                   at={(axis description cs:0.5,-0.07)},anchor=north, font=\scriptsize},
                   %ylabel=BLEU,
                   xtick={-3,-1,1,3,...,10},
                   minor x tick num=1,
                   cycle list name=mainpalette,
                   ]

\addplot coordinates {(-5, \wue) (22, \wue)};
\addlegendentry{Offline}

\addplot table [y=BLEU4-b1,x=AL-b1]{\results/unidir_transformer/trained_path_kinf.dat};
\addlegendentryexpanded{$\ktr=\infty$}

\addplot table [y=BLEU4-b1,x=AL-b1]{\results/unidir_transformer/trained_path_k7.dat};
\addlegendentry{$\ktr=7$}

\addplot table [y=BLEU4-b1,x=AL-b1]{\results/unidir_transformer/trained_path_k1.dat};
\addlegendentry{$\ktr=1$}

\addplot table [y=BLEU4-b1,x=AL-b1]{\results/unidir_transformer/keval.dat};
\addlegendentry{$\ktr=\kev$}

\addplot table [y=BLEU4-b1,x=AL-b1]{\results/unidir_transformer/1tolx.dat};
\addlegendentry{$\ktr\in[1..\lx]$}

\legend{}
\end{groupplot}
\end{tikzpicture}
    \caption{Vi$\sto$En Transformer}\label{fig:vien-mt-path}
\end{subfigure}%
\vspace{-5pt}
\caption{IWSLT'14 De$\leftrightarrow$En and IWSLT'15 En$\leftrightarrow$Vi: wait-$k$ online decoding with Pervasive Attention (top) and Transformer (bottom), both with unidirectional encoder. Each curve represents  a model  trained on a single decoding path, evaluated  with $\kev\in \{1, 3, 5, 7, 9\}$. 
    Offline models have an average lagging of 22.97, 22.21, 21.08 and 26.56 on De$\sto$En, En$\sto$De, En$\sto$Vi and Vi$\sto$En, respectively.
}\label{fig:iwslt-main}
\vspace{-5pt}
\end{figure*}
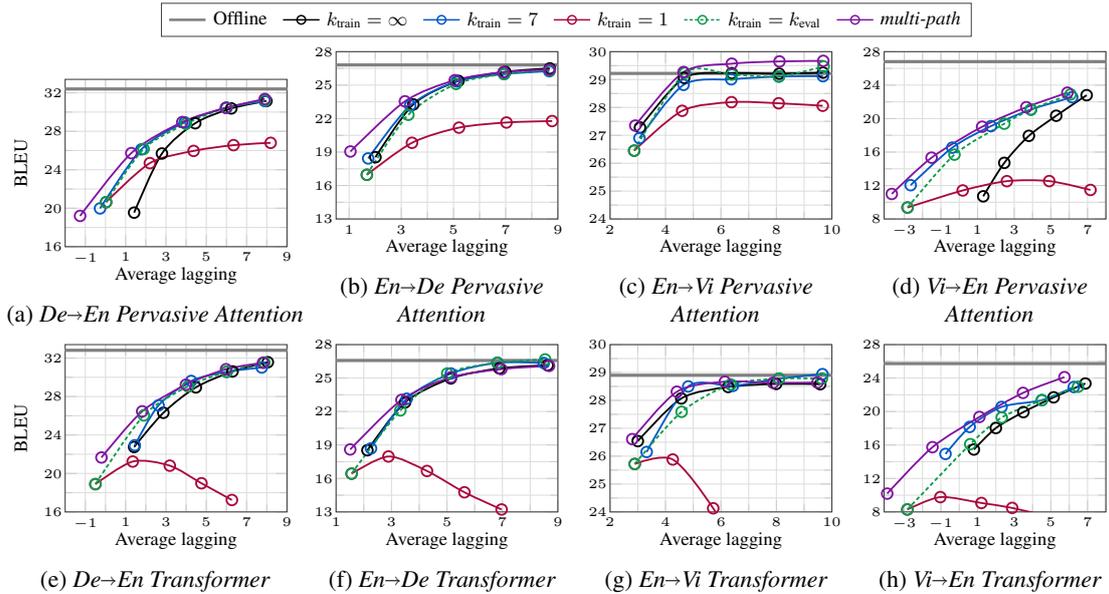

\begin{figure*}[ht]
\centering
\begin{subfigure}{.29\linewidth}
    \def\results{data/wmt15_deen}
\begin{tikzpicture}
\pgfplotsset{
    compat=1.11,
    legend image code/.code={
    \draw[mark repeat=2,mark phase=2, mark size=1.2pt]
    plot coordinates {
    (0cm,0cm)
    (0.15cm,0cm)        %% default is (0.3cm,0cm)
    (0.3cm,0cm)         %% default is (0.6cm,0cm)
    };}}

\pgfmathsetmacro{\avleng}{26.96}
\pgfmathsetmacro{\Basewue}{31.14}
\pgfmathsetmacro{\Bigwue}{31.46}
\pgfmathsetmacro{\MPAwue}{27.91}
\begin{groupplot}[
    group style={%
        group size=1 by 1,
        horizontal sep=0pt
    },
    ymin=18, ymax=32,
    legend columns=1,
    legend style={at={(-0.03,1.05)},
    anchor=south west, 
    font=\tiny, legend cell align=left, font=\tiny, inner sep=0.5pt},
    ytick={18,20,...,32},
    minor y tick num=1,
    tick label style={font=\tiny},
    label style={font=\scriptsize},
    y tick label style={font=\tiny, xshift=2pt},
    x tick label style={font=\tiny, yshift=1pt},
   ]
\nextgroupplot[xmin=-0.75,xmax=18,
               %axis y line=left,
               height=4.0cm,
               width=5.5cm, grid=both,
               y axis line style=-,
               xlabel=\csname AL\endcsname, 
               x label style={%
               at={(axis description cs:0.5,-0.07)},anchor=north, font=\scriptsize},
               minor x tick num=1, 
               xtick={-2,2,...,28},
               ylabel=BLEU,
               y label style={font=\scriptsize},
               legend to name=WMTbig,
               ]

\addplot +[mark=none, line width=1.2pt, gray] coordinates {(-1, \Bigwue) (22, \Bigwue)};
\addlegendentry{Offline}

\addplot[smooth, black, mark=o, mark size=2pt, line width=.7pt]
table [y=BLEU4-b1,x=AL-b1]{\results/unidir_transformer/trained_path_kinf_big.dat};
\addlegendentry{$\ktr{=}\infty$}

\addplot[smooth, p1blue, mark=o, mark size=2.pt, line width=.7pt]
table [y=BLEU4-b1,x=AL-b1]{\results/unidir_transformer/big_trained_path_k7.dat};
\addlegendentry{$\ktr{=}7$}

\addplot[p1green, mark=o, mark size=2pt, line width=0.9pt, densely dotted] 
table [y=BLEU4-b1,x=AL-b1]{\results/unidir_transformer/keval_big.dat};
\addlegendentry[align=left]{$\ktr{=}\kev$}

\addplot[smooth, p1purple, mark=o, mark size=2.pt, line width=0.7pt] 
table [y=BLEU4-b1,x=AL-b1]{\results/unidir_transformer/trained_1tolx_big.dat};
\addlegendentry{\emph{multi-path}}

\end{groupplot}
\node[anchor=south east, scale=0.85] at (rel axis cs: 1.004,-0.03) {\pgfplotslegendfromname{WMTbig}};

\end{tikzpicture}
    \caption{Our TF big models}\label{fig:wmt1}
\end{subfigure}%
\begin{subfigure}{.27\linewidth}
    \def\results{data/wmt15_deen}
\begin{tikzpicture}
\pgfplotsset{
    compat=1.11,
    legend image code/.code={
    \draw[mark repeat=2,mark phase=2, mark size=1.2pt]
    plot coordinates {
    (0cm,0cm)
    (0.15cm,0cm)        %% default is (0.3cm,0cm)
    (0.3cm,0cm)         %% default is (0.6cm,0cm)
    };}}
\pgfmathsetmacro{\avleng}{26.96}
\pgfmathsetmacro{\MTwue}{31.14}
\pgfmathsetmacro{\MPAwue}{27.91}

\pgfmathsetmacro{\Basewue}{31.14}
\pgfmathsetmacro{\Bigwue}{31.46}

\begin{groupplot}[
    group style={%
        group size=2 by 1,
        horizontal sep=0pt
    },
    ymin=18, ymax=32,
    legend columns=1,
    legend style={at={(-0.05,1.05)},
    anchor=south west, inner sep=0.5pt, 
    font=\tiny, legend cell align=left},
    ytick={18,20,...,32},
    minor y tick num=1,
    tick label style={font=\tiny},
    label style={font=\scriptsize},
    y tick label style={font=\tiny, xshift=2pt},
    x tick label style={font=\tiny, yshift=1pt},
   ]
\nextgroupplot[xmin=-0.75,xmax=18,
           %axis y line=left,
           height=4.0cm, width=5.5cm, 
           grid=both, y axis line style=-,
           xlabel=\csname AL\endcsname, 
           x label style={at={(axis description cs:0.5,-0.07)},anchor=north, font=\scriptsize},
           minor x tick num=1, xtick={-2,2,...,28},
           %ylabel=BLEU,
           legend to name=WMTsota1,
           ]

\addplot +[mark=none, line width=1.2pt, gray] coordinates {(-1, \Basewue) (22, \Basewue)};
\addlegendentry{Offline}

\addplot[smooth, black, mark=o, mark size=2pt, line width=.7pt]
table [y=BLEU4-b1,x=AL-b1]{\results/unidir_transformer/trained_path_kinf.dat};
\addlegendentry{$\ktr{=}\infty$}

\addplot[p1green, mark=o, mark size=1.8pt, line width=0.9pt, densely dotted] 
table [y=BLEU4-b1,x=AL-b1]{\results/unidir_transformer/keval.dat};
\addlegendentry[align=left]{$\ktr{=}\kev$}

\addplot[p1purple, mark=o, mark size=1.8pt, line width=0.7pt] 
table [y=BLEU4-b1,x=AL-b1]{\results/unidir_transformer/trained_1tolx.dat};
\addlegendentry{\emph{multi-path}}

\addplot[p1red, mark=o, mark size=1.8pt, line width=0.9pt, densely dotted] 
table [y=BLEU4-b1,x=AL-b1]{\results/stacl_waitk.dat};
\addlegendentry[align=left]{\emph{STACL} $\ktr{=}\kev$~\cite{Ma19acl}}

\addplot[smooth, gray, mark=o, mark size=1.8pt, line width=0.7pt] 
table [y=BLEU4-b1,x=AL-b1]{\results/simpler_and_faster_dynamic.dat};
\addlegendentry[align=left]{\emph{SL}~\cite{Zheng19emnlp}}

\end{groupplot}
\node[anchor=south east, scale=.75] at (rel axis cs: 1.003,-0.03) {\pgfplotslegendfromname{WMTsota1}};

\end{tikzpicture}
    \caption{SoTA comparison, TF base}\label{fig:wmt2}
\end{subfigure}%
\begin{subfigure}{.27\linewidth}
    \def\results{data/wmt15_deen}
\begin{tikzpicture}
\pgfplotsset{
    compat=1.11,
    legend image code/.code={
    \draw[mark repeat=2,mark phase=2, mark size=1.2pt]
    plot coordinates {
    (0cm,0cm)
    (0.15cm,0cm)        %% default is (0.3cm,0cm)
    (0.3cm,0cm)         %% default is (0.6cm,0cm)
    };}}

\pgfmathsetmacro{\avleng}{26.96}
\pgfmathsetmacro{\MTwue}{31.14}
\pgfmathsetmacro{\MPAwue}{27.91}

\pgfmathsetmacro{\Basewue}{31.14}
\pgfmathsetmacro{\Bigwue}{31.46}

\begin{groupplot}[
    group style={%
        group size=2 by 1,
        horizontal sep=0pt
    },
    ymin=18, ymax=32,
    legend columns=1,
    legend style={at={(-0.05,1.05)},
    anchor=south west, inner sep=0.5pt, 
    font=\tiny, legend cell align=left},
    ytick={18,20,...,32},
    minor y tick num=1,
    tick label style={font=\tiny},
    label style={font=\scriptsize},
    y tick label style={font=\tiny, xshift=2pt},
    x tick label style={font=\tiny, yshift=1pt},
   ]
\nextgroupplot[xmin=-0.75,xmax=16,
               %axis y line=left,
               height=4.0cm,
               width=5.5cm, 
               grid=both,
               y axis line style=-,
               xlabel=\csname AL\endcsname, 
               x label style={at={(axis description cs:0.5,-0.07)},anchor=north, font=\scriptsize},
               minor x tick num=1,
               xtick={-2,2,...,28},
               %ylabel=BLEU,
               legend to name=WMTsota2,
               ]

\addplot[p1purple, mark=o, mark size=2pt, line width=0.7pt] 
table [y=BLEU4-b1,x=AL-b1]{\results/unidir_transformer/trained_1tolx_big.dat};
\addlegendentry{\emph{multi-path}}

\addplot[smooth, p1azure, mark=x, mark size=2pt, line width=0.7pt, densely dotted]
table [y=BLEU4-b1,x=AL-b1]{\results/mocha.dat};
\addlegendentry[align=left]{\emph{MoChA}~\cite{Chiu18iclr}}

\addplot[smooth, p1blue,mark=x, mark size=2pt, line width=0.7pt, densely dotted]
table [y=BLEU4-b1,x=AL-b1]{\results/milk.dat};
\addlegendentry[align=left]{\emph{MILk}~\cite{Arivazhagan19acl}}

\addplot[smooth, p1orange, mark=x, mark size=2pt, line width=0.7pt, densely dotted]
table [y=BLEU4-b1,x=AL-b1]{\results/mma_h.dat};
\addlegendentry[align=left]{\emph{MMA-H}~\cite{Ma20iclr}}

\addplot[smooth, p1red, mark=x, mark size=2pt, line width=0.7pt, densely dotted]
table [y=BLEU4-b1,x=AL-b1]{\results/mma_il.dat};
\addlegendentry[align=left]{\emph{MMA-IL}~\cite{Ma20iclr}}

\end{groupplot}
\node[anchor=south east] at (rel axis cs: 1.02,-0.03) {\pgfplotslegendfromname{WMTsota2}};

\end{tikzpicture}
    \caption{SoTA comparison, TF big}\label{fig:wmt3}
\end{subfigure}%
\vspace{-5pt}
\caption{
Evaluation of our models on WMT'15 De$\sto$En, and comparison to the state of the art (SoTA).
Offline models have an average lagging of 26.96.
Note that in panel (b) STACL $\ktr=\kev$ shows the results of \cite{Zheng19emnlp} where an end-of-sequence marker was added to the source to improve over \cite{Ma19acl}.
}
\label{fig:wmt}
\vspace{-5pt}
\end{figure*}
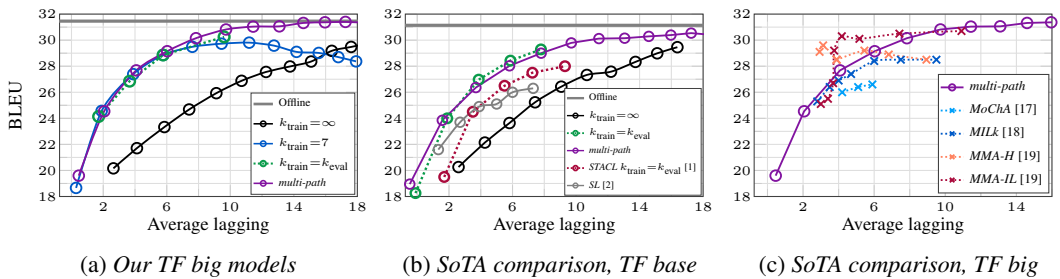

\fig{iwslt-main} presents the performance of models trained for a single wait-$k$ decoding path, with  $\ktr\in\{1, 7, \infty\}$. 
Each trained model is represented by a curve, by evaluating it across different wait-$k$ decoding paths $\kev\in\{1,3,5,7,9\}$.

Initial experiments with both architectures across the four IWSLT tasks showed that models trained on wait-7 generalize well on  other evaluation paths. Thus, unlike \cite{Ma19acl}, we note that we can train a single model to use in different latency regimes, \ie we do not  achieve better BLEU scores when training with  $\kev=\ktr$.
This generalization to other wait-$k$ paths is notably stronger with pervasive attention (PA) models. 
Where TF models drop in performance far from the training path (\eg $\ktr=1$ and $\kev=9$), the PA models continue to improve  for larger  $\kev$.
Overall, for tasks where PA performs better offline the model consistently outperforms TF online and vice-versa.
It is worth noting that for some translation directions, we can outperform the offline model's performance at a considerably lower latency. This is in particular the case for En$\sto$Vi where
the online PA with  an AL of 4.65 matches the  performance of the offline model with an AL of 21.08,  see \fig{envi-mpa-path}.

\mypar{Joint training on multiple paths}
We found that training on a particular wait-$k$ path can generalize well to other paths. 
To avoid tuning $\ktr$ to find the optimal path for each specific task, we consider jointly optimizing on multiple paths.
Results in \fig{iwslt-main} show that this joint optimization, for both architectures and on two datasets, manages to achieve comparable or better results than training on a single manually selected   path.

\mypar{Experiments on the WMT15 De$\sto$En}
On WMT15 De$\sto$En we experiment with transformer \emph{base} (comparable to \cite{Ma19acl, Zheng19emnlp} and \emph{big} (comparable to \cite{Arivazhagan19acl, Ma20iclr}).\footnote{\emph{MILk} is based on RNMT+ which outperforms TF \emph{big} offline \cite{Chen18acl}.}
In \fig{wmt1} we observe, as for IWSLT, that jointly training on multiple wait-$k$ paths outperforms training on a single path where the performance drops as we move away from $\ktr$.
The advantage of joint training with unidirectional encoders is confirmed in \fig{wmt2} when comparing our results to \emph{STACL} which trains separate bidirectional models for each decoding path with  $\ktr=\kev$. 
Our models also outperform SL~\cite{Zheng19emnlp} that optimize dynamic agents with imitation learning (IL).

Both our \emph{base} and \emph{big} \emph{multi-path} models match or improve the performance of the dynamic \emph{MILk}~\cite{Arivazhagan19acl} that requires training for each latency regime (each mark in the dotted curves is a different model) whereas our wait-$k$ model is simply evaluated with different values of $\kev$.
The more recent \emph{MMA-H} and \emph{MMA-IL}~\cite{Ma20iclr} adapting \emph{MoChA} and \emph{MILk} for Transformer models outperform \emph{wait-$k$} models for $\textrm{AL}<6$,  but fail to optimize a medium lagging model.

\section{Conclusion}\label{sec:concl}
\vspace{0pt}
In this paper, we demonstrated that unidirectional encoders for online MT achieve better translation qualities than bidirectional ones, with faster training and decoding. 
Moreover, we introduced joint training for \emph{wait-$k$} decoders addressing the need to train a different model for each lagging value. 
Our  models are trained end-to-end and, unlike conventional \emph{wait-$k$}, can operate across the full spectrum of lagging with the quality increasing with the value of $k$.
In low-resource settings, we found Pervasive Attention models to be  competitive with Transformers for online translation.
Our  wait-$k$ models are state-of-the-art among deterministic  online translation strategies, and  provide a strong baseline  for   simultaneous translation with dynamic decoding.

\bibliographystyle{IEEEtran}
\bibliography{references}

\end{document}